\documentclass[10pt,twocolumn,letterpaper]{article}

\usepackage{iccv}
\usepackage{times}
\usepackage{epsfig}
\usepackage{graphicx}
\usepackage{amsmath}
\usepackage{amssymb}

\usepackage[pagebackref=true,breaklinks=true,letterpaper=true,colorlinks,bookmarks=false]{hyperref}

\iccvfinalcopy 

\def\iccvPaperID{3374} 

\ificcvfinal\fi

\newcommand{\Skip}[1]{}

\newcommand{\WJ}[1] {
	\textcolor{magenta}{\bfseries{WJ: {#1}}}
}

\usepackage{booktabs}
\usepackage{multirow}
\usepackage{color}
\usepackage{colortbl}
\usepackage{makecell}
\usepackage{authblk}

\begin{document}

\title{Towards Content-based Pixel Retrieval in Revisited Oxford and Paris}

\Skip{
\author{Guoyuan An\\
KAIST\\
South Korea\\
{\tt\small anguoyuan@kaist.ac.kr}
\and
Woo Jae Kim\\
KAIST\\
South Korea\\
{\tt\small wkim97@kaist.ac.kr}
}}

\author[1]{\bf Guoyuan An}
\author[1]{\bf Woo Jae Kim}
\author[1]{\bf Saelyne Yang}
\author[3]{\bf Rong Li}
\author[2,3]{\bf Yuchi Huo}
\author[1]{\bf Sung-Eui Yoon}

\affil[1]{School of Computing, KAIST}
\affil[2]{ State Key Lab of CAD\&CG, Zhejiang University}
\affil[3]{Zhejiang Lab}

\maketitle
\ificcvfinal\thispagestyle{empty}\fi

\begin{abstract}
   This paper introduces the first two pixel retrieval benchmarks. Pixel retrieval is segmented instance retrieval. Like semantic segmentation extends classification to the pixel level, pixel retrieval is an extension of image retrieval and offers information about which pixels are related to the query object. In addition to retrieving images for the given query, it helps users quickly identify the query object in true positive images and exclude false positive images by denoting the correlated pixels. Our user study results show pixel-level annotation can significantly improve the user experience. 
   Compared with semantic and instance segmentation, pixel retrieval requires a fine-grained recognition capability for variable-granularity targets. To this end, we propose pixel retrieval benchmarks named PROxford and PRParis, which are based on the widely used image retrieval datasets, ROxford and RParis. Three professional annotators label 5,942 images with two rounds of double-checking and refinement. Furthermore, we conduct extensive experiments and analysis on the SOTA methods in image search, image matching, detection, segmentation, and dense matching using our pixel retrieval benchmarks. Results show that the pixel retrieval task is challenging to these approaches and distinctive from existing problems, suggesting that further research can advance the content-based pixel-retrieval and thus user search experience. The datasets can be downloaded from \href{https://github.com/anguoyuan/Pixel_retrieval-Segmented_instance_retrieval}{this link}. 
\end{abstract}

\section{Introduction}
\label{sec:Introduction}

\begin{figure}[t]
    \centering
    \includegraphics[scale=0.4]{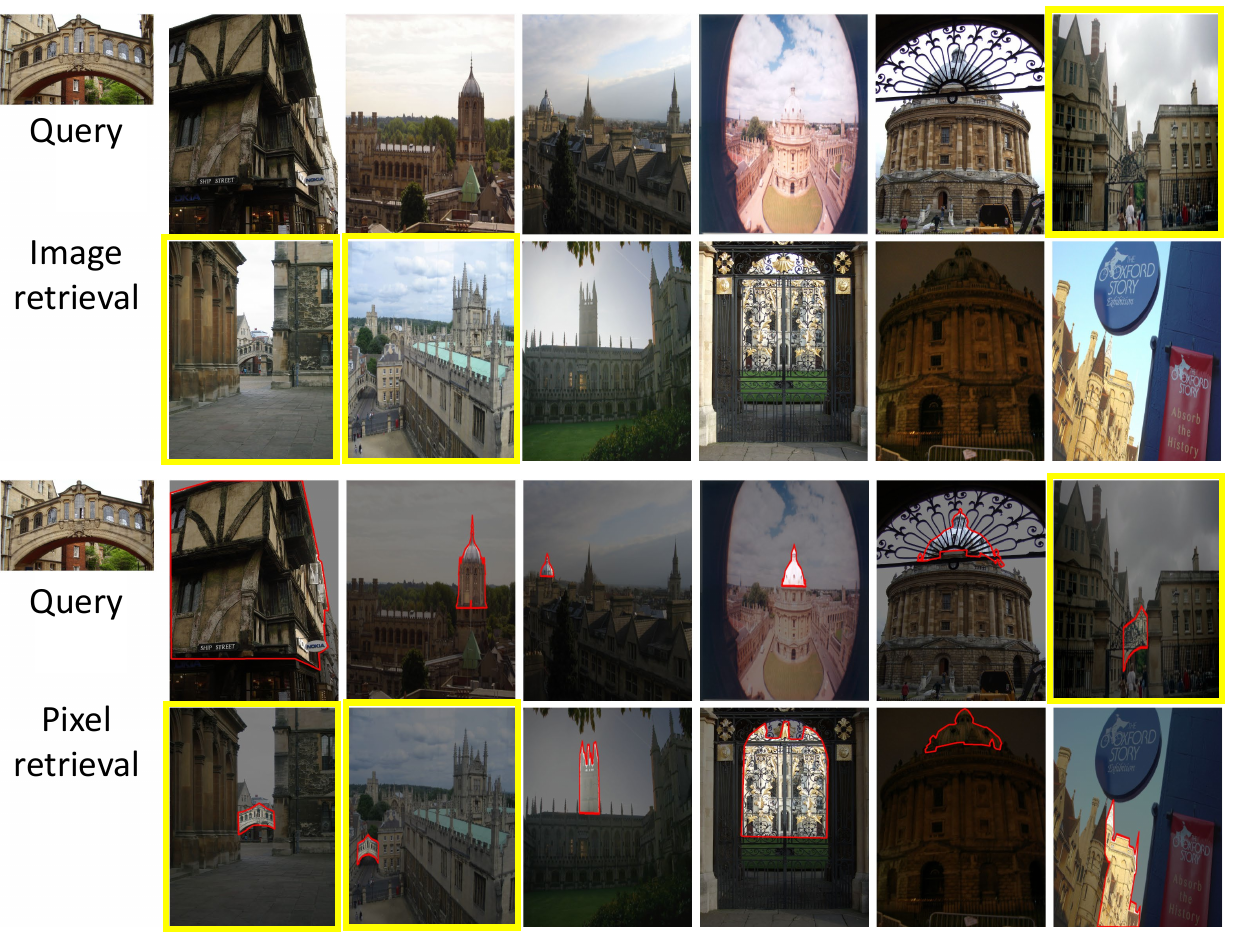}
    \caption{Example scenarios of image retrieval and pixel retrieval for the same query image. Pixel retrieval offers pixel-level annotation (red outlines) on the target object. Our user study shows that pixel retrieval can significantly improve the user experience (Sec.~\ref{sec:Applications}). Yellow boxes in the searched results indicate
   the ground truth ones.
  You can check our user study from \href{https://fascinating-marzipan-a99b4c.netlify.app/bwds}{\textcolor{red}{this link}}. To start the user study, please enter any character into the ``unique Prolific ID'' blank. }
    \label{fig:user_study}
\end{figure}

Image retrieval is a long-standing and fundamental computer vision task and has achieved remarkable advances. However, because the retrieved ranking list contains false positive images and the true positive images contain complex co-occurring backgrounds, users may be difficult to identify the query object from the ranking list. In this paper, we execute a user study and show that providing pixel-level annotations can help users better understand the retrieved results. Therefore, this paper introduces the pixel retrieval task and its first benchmarks. Pixel retrieval is defined as searching pixels that depict the query object from the database. More specifically, it requires the machine to recognize, localize, and segment the query object in database images in run time, as shown in Figure~\ref{fig:user_study}.

Similar to semantic segmentation, which works as an extension of classification and provides pixel-level category information to the machines, pixel retrieval is an extension of image retrieval. However, pixel retrieval differs from existing semantic segmentation~\cite{he2017mask,zhou2017scene,liu2018path} in two aspects: the fine-grained particular instance recognition and the variable-granularity recognition. 

On the one hand, pixel retrieval asks the machine to consider the fine-grained information to segment the same instance with the query, \textit{e.g.}, segment the particular query building in the street figures that contain many similar buildings. This is different from existing semantic segmentation~\cite{he2017mask} and instance segmentation~\cite{liu2018path,zhou2017scene}. Semantic segmentation only requires the category level information, \textit{e.g.}, to segment all the buildings in the street figures. On top of semantic segmentation, instance segmentation additionally requires demarcating individual instances, \textit{e.g.}, segmenting all the buildings and giving the boundary of each building separately. However, instance segmentation does not distinguish the differences among the buildings~\cite{zhou2017scene,liu2018path,bolya2019yolact}.

On the other hand, pixel retrieval requires adjusting the recognition granularity as needed. The query image can be the whole building or only a part of the building. The search engine should understand the intention of the query and adjust the segmentation granularity in demand. This differs from existing segmentation benchmarks~\cite{zhou2017scene,cordts2016cityscapes,lin2014microsoft,everingham2010pascal,geiger2012we}, where the recognition granularity is fixed in advance. Therefore, the pixel retrieval task is supplementary to semantic and instance segmentation by considering the recognition and segmentation featured with fine-grained and variable-granularity properties, which are also fundamental visual abilities of humans. 

\Skip{
Pixel retrieval is useful in many practical applications. Modern image retrieval techniques focus on improving
the image-level ranking performance of the hard cases~\cite{cao2020unifying,an2021hypergraph,radenovic2018revisiting}. However, because of the difficulty of the hard case, users may not be conscious that it is truly positive in a short time, even though it is at the top of the ranking list.
\WJ{Again, the connection between trends of recent ranking techniques and user experience is unclear.
Can we explain how pixel retrieval can improve user experience? 
e.g., By highlighting the retrieved pixels on the searched images, we can guide the users to locate regions of interest more easily in such cases.}
Our user study results show pixel-level annotation can significantly improve the user experience (Section~\ref{sec:Applications}).

Besides the web search, improving the pixel retrieval performance helps the image level ranking techniques. Pixel retrieval can also be useful in medical diagnosis~\cite{yang2014parallel}, geographical information systems~\cite{zhu2021vigor}, and image matting~\cite{wei2021improved,xu2017deep,zhang2019late} (Section ~\ref{sec:Applications}). }

In order to promote the study of pixel retrieval, we create the pixel retrieval benchmarks Pixel-Revisited-Oxford (PROxford) and Pixel-Revisited-Paris (PRParis) on top of the famous image retrieval benchmarks Revisited-Oxford (ROxford) and Revisited-Paris (RParis)~\cite{philbin2007object, philbin2008lost,radenovic2018revisiting}. There are three reasons to use ROxford and RParis as our base benchmarks. Firstly, they are notoriously difficult and can better reflect the search engines' performance. Secondly, each query in these datasets has up to hundreds of positive images, so they are suitable for evaluating the fine-grained recognition ability. Thirdly, every positive image is guaranteed to be identifiable by people without considering any contextual visual information~\cite{radenovic2018revisiting}. 

We provide the segmentation labels to a total of 5,942 images in ROxford and RParis. 
To ensure the label quality, three professional annotators independently label the query-index pairs and then refine and check the labels. The annotators are aged between 26 to 32 years old and have worked full-time on annotation for over two years. We then design new metrics, mAP@50:5:95, and mAP, to evaluate the pixel retrieval performance (Section~\ref{sec:Dataset}).

We provide an extensive comparison of State-Of-The-Art (SOTA) methods in related fields, including image search, detection, segmentation, and dense matching with our benchmarks. We have some interesting findings from the experiment. For example, we find the SOTA spatial verification methods~\cite{cao2020unifying,Noh_2017_ICCV} give a high inlier number to some true query-index pairs but match the wrong regions. We find the dense and pixel-level approaches~\cite{min2021hypercorrelation, truong2021warp} helpful for the pixel retrieval task. Most importantly, our results show that pixel retrieval is difficult and further research is needed for advancing the user experience on the content-based search task. 

Our contributions are as follows: 
\begin{itemize}
    \item We introduced the pixel retrieval task and provided the first two landmark pixel retrieval benchmarks, PROxford and PRParis. Three professional annotators labeled, refined, and checked the labels.
    \item We executed the user study and showed that the pixel level annotation could significantly improve user experience.
    \item We performed extensive experiments with SOTA methods in image search, detection, segmentation, and dense matching. Our experiment results can be used as the baselines for future study.
\end{itemize}

\section{Content-based pixel retrieval}
\label{sec:Dataset}

\subsection{Why Revisited Oxford and Paris?}
We design the first content-based pixel retrieval benchmarks, PROxford and PRParis, directly on top of the famous image retrieval benchmarks Revisited-Oxford (ROxford) and Revisited-Paris (RParis)~\cite{philbin2007object, philbin2008lost,radenovic2018revisiting}. Oxford~\cite{philbin2007object} and Paris~\cite{philbin2008lost} are introduced by Philbin~\etal in 2007 and 2008, respectively. Their images are obtained from Flickr by searching text tags for famous landmarks in Oxford University and Paris. Radenovic~\etal~\cite{radenovic2018revisiting} refined the annotations and updated more difficult queries for them in 2018; the refined datasets are called ROxford and RParis. 

We choose ROxford and RParis because they are among the most popular image retrieval benchmarks. Many well-known image retrieval methods are evaluated on them, from the traditional methods like RootSIFT~\cite{arandjelovic2012three}, VLAD~\cite{jegou2010aggregating}, and ASMK~\cite{tolias2015particular}, to the recent deep learning based methods like R-MAC~\cite{tolias2015particular}, GeM~\cite{radenovic2018fine}, and DELF~\cite{Noh_2017_ICCV}.

These datasets are the ideal data sources for our pixel-retrieval, thanks to several properties. Firstly, compared to other famous datasets like image matching Phototourism~\cite{Jin2020} and dense matching Megadepth~\cite{li2018megadepth}, the positive image pairs in ROxford and RParis have severe viewpoint changes, occlusions, and illumination changes. The new queries added by Radenovic~\etal~\cite{radenovic2018revisiting} have cropped regions that cause extreme zooms with the positive database images. These properties make the ROxford and RParis notoriously difficult. Secondly, each query image contains up to hundreds of positive database images, while other datasets,
such as UKBench~\cite{nister2006scalable} and Holiday~\cite{jegou2008hamming}, only have 4 to 5 positive images for each query. A large amount of challenging positive images are suitable for evaluating fine-grained recognition ability.

The Google Landmark Dataset (GLD)~\cite{weyand2020google} encompasses more landmarks than ROxford and RParis. However, ROxford and RParis outshine GLD in labeling quality. Notably, they stand as distinct benchmarks for contrasting machine and human recognition prowess.

It is known that people cannot easily recognize an object if it changes its pose significantly~\cite{ponce2006dataset}, but we do not know where the limit is. ROxford and RParis are \textbf{the only existing datasets that can reflect the human ability to identify objects} in the landmark domain to the best of our knowledge. 
Every positive image in ROxford and RParis is checked by five annotators independently based on the image appearance, and all the unclear cases are excluded~\cite{radenovic2018revisiting}. This kind of annotation has two benefits. Firstly, although these benchmarks are difficult, the positive images are guaranteed to be identifiable by people without considering any contextual visual information~\cite{radenovic2018revisiting}. This shows the possibility of enabling the machine to recognize these positive images by only analyzing the visual clue in the given query-index image pair. Secondly, these datasets can be used to compare human and machine recognition performance; human-level recognition performance should identify all the positive images. 
Although the classification performance (the top 5 accuracy) of machines on ImageNet has surpassed that of humans~\cite{russakovsky2015imagenet}, the SOTA identification ability about the first-seen objects in ROxford and RParis is still far from human-level~\cite{lee2022correlation,an2021hypergraph,cao2020unifying}.

\subsection{From image retrieval to pixel retrieval}
In a similar spirit that semantic segmentation works as an extension of classification and provides pixel-level category information to the machines, pixel retrieval is an extension of image retrieval. It offers information about which pixels or regions are related to the query object. This task is very helpful when only a small region of the positive image corresponds to the query. Such situations frequently happen in many image retrieval applications, such as web search~\cite{radenovic2018revisiting, lampert2009detecting, lin2010local}, medical image analysis~\cite{mehta2009content,breznik2022cross,yang2014parallel}, geographical information systems~\cite{zhang2022dataset,zhu2021vigor, shi2022beyond}, and so on. We discuss the related applications in Section~\ref{sec:Applications}. Distinguishing and segmenting the first-seen objects is also one basic function of human vision system~\cite{shrager2006intact}; it is meaningful to understand and automate this ability. \Skip{can trigger many applications.}

Some previous works also noticed the importance of localizing the query object in the searched image. They have tried to combine image search and object localization~\cite{lampert2009detecting,lin2010local,shen2013spatially}. However, due to the lack of a challenging pixel retrieval benchmark, they show only the qualitative result instead of the quantitative performance. Pixel-level labeling and quality assurance are arduous. In this work, 5,942 images are labeled, refined, and checked by three professional annotators. We hope this benchmark can boost and encourage future research on pixel-level retrieval.

We also compare our pixel-retrieval benchmark with segmentation, image matching, and dense matching benchmarks in the supplementary material.

\begin{figure}
    \centering
    \includegraphics[width=0.9\linewidth]{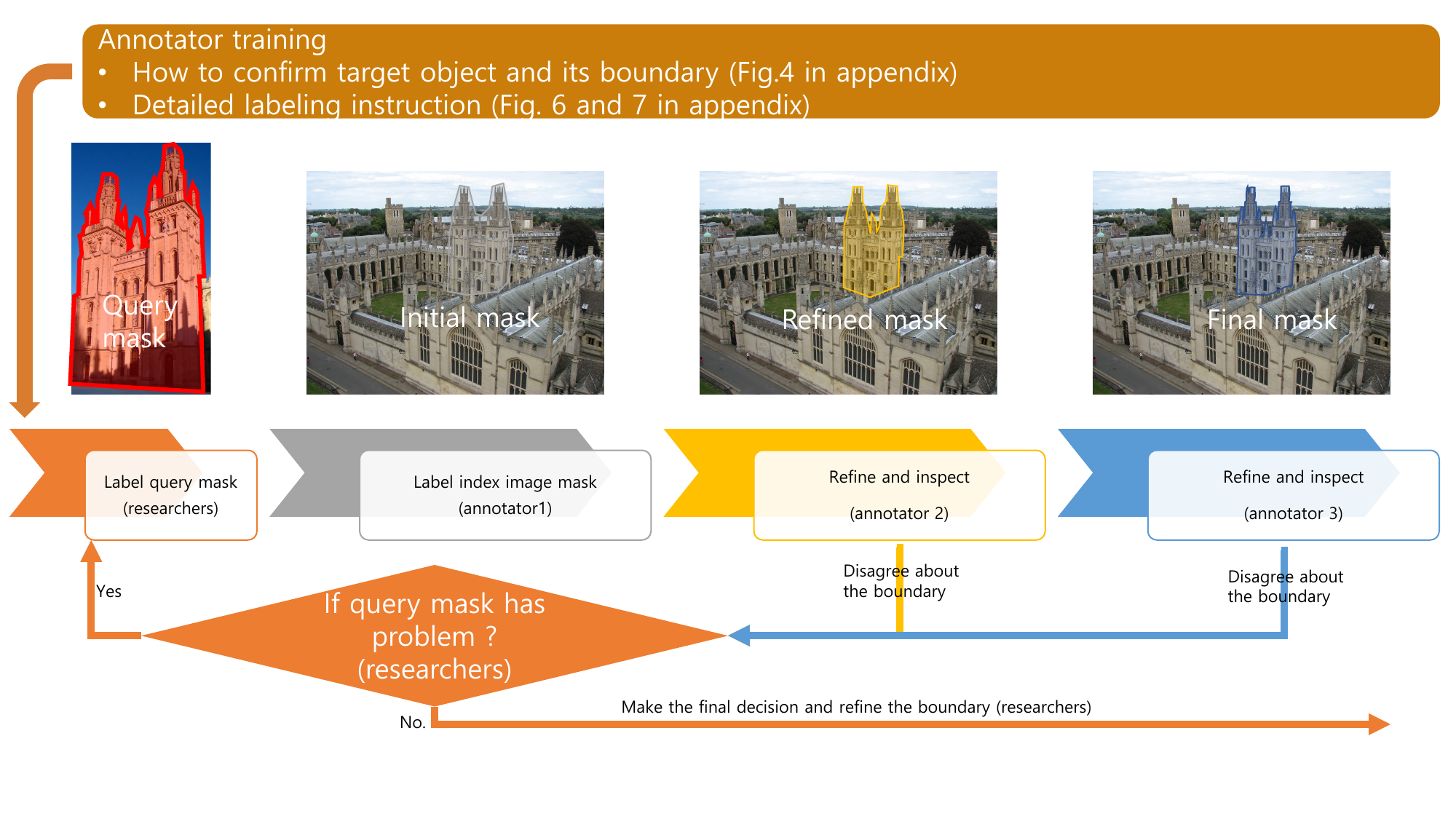}
    \vspace{-10pt}
    \caption{Labeling process (please zoom in for details).}
    \label{fig:label}
\end{figure}

\subsection{Pixel-level annotation}

\noindent{\bf Images to annotate.}
ROxford and RParis each contains 70 queries. The 70 queries are divided into 26 and 25 query groups in ROxford and RParis, respectively, based on the visual similarity; queries in the same query group share the same ground truth index image list. 
There are total 1,985 and 3,957 images to annotate for our PROxford and PRParis, respectively.

\noindent{\bf Mask annotation.}
Figure~\ref{fig:label} shows our labeling process. Researchers with a computer vision background first annotate the target object in each query image. 
Each annotator for our new benchmark observes all the queries with masks in a query group and labels the segmentation mask for the images in the ground-truth list. Annotators are asked to identify the query object in the labeling image first and then label all the pixels depicting the target object. We show the query masks and the labeling instruction details in the supplementary materials. 

\Skip{
We give the annotators the following instructions, with more detailed instructions listed in the supplementary materials.
\begin{itemize}
    \item Identify the query object in the labeling image firstly. Please zoom in and zoom out the query and the labeling image for observation if needed. If no query object is found in the labeling image in 3 minutes, report and skip this image.
    \item Label all the pixels belonging to the identified object. Due to the viewpoint change, not every correct pixel in the labeling image has a corresponding pixel in the query image. Label the pixel as long as it is depicting the target object.
    \item The target object may connect to or is a part of another object. Please judge the intention of the query example.
    \item A part of the target object may be occluded by objects like a fence and window. Removing all the occlusion pixels is not necessary. However, there should be only one clear salient object in the mask region. Remove all the pixels that violate this rule.
\end{itemize}
}

\noindent{\bf Objectivity.}
To ensure the pixel retrieval task and our benchmark are objectively defined, we adopt two approaches. Firstly, we use query masks to distinctly identify the target objects and segregate them from the background (\textit{e.g.,} the sky), occlusions (\textit{e.g.,} other buildings), and the remaining part of the same building if the object is only a small part of it. These masks guide the removal of background and indicate the query boundary. Secondly, by examining the query with masks, our annotators reach a consensus on the target object and its boundary, thereby avoiding disagreement about our query intention. This consensus-based approach is a common method for reducing subjectivity in recognition tasks; it is also employed in the original ROxford and RParis benchmarks, where voting is used to determine the final ground truth for each query~\cite{radenovic2018revisiting}. 

We retain small-sized occlusion objects, like windows and fences, during annotation. While this may involve subjective judgments regarding what qualifies as a small-sized occlusion, it is worth noting that well-known semantic segmentation datasets like VOC~\cite{everingham2010pascal} and COCO~\cite{lin2014microsoft} also involve subjective elements, such as identifying objects on a table as a table or the background behind the bike wheel as a bike. Such subjectivities are inevitable, given the difficulty of removing them. Nonetheless, they do not diminish the usefulness of benchmarks as reliable metrics for evaluating state-of-the-art methods. 
We include in the supplementary materials our mask rules, all the queries with masks, and our consensus checking.

\noindent{\bf Quality assurance.}
To improve the annotation quality, every query-index image pair labeling is performed by three
professional annotators following the three steps: 1) anno-
tate; 2) refine + inspect; 3) refine + insp, as shown in Figure~\ref{fig:label}. 
\Skip{The 3 steps are sequentially did by three professional annotators. }
The three annotators are aged between 26 to 32 years old and have worked on annotation full-time for over 2 years. Their works have been qualified in many annotation projects. 

\subsection{Evaluation metrics}

\noindent{\bf Pixel retrieval from the database.}
Pixel retrieval aims to search all the pixels depicting the query object from the large scale database images. An ideal pixel retrieval algorithm should achieve the image ranking, reranking, localization, and segmentation simultaneously. To the best of our knowledge, there is no existing pixel retrieval metric yet. Detection and segmentation tasks usually use mIoU and mAP@50:5:95 as the standard measurement~\cite{ren2015faster}. Image retrieval methods commonly use mAP as the metric~\cite{radenovic2018revisiting}. We combine them to evaluate the ranking, localization, and segmentation performance in pixel retrieval. Each ground-truth image in the ranking list is treated as a true-positive (TP), only if its detection or segmentation Intersection over Union (IoU) is larger than a threshold $n$. The other process of calculating AP and mAP follows the traditional image search mAP. Note that the mAP calculation methods in image search and traditional segmentation~\cite{everingham2010pascal} are different; image search focuses more on ranking. Similar to detection and segmentation fields, the threshold $n$ is set from 0.5 to 0.95, with step 0.05. The average of scores under these thresholds are the final metric mAP@50:5:95. It is desirable to report both detection and segmentation mAP@50:5:95 for the methods that can generate pixel-level results; high segmentation performance does not necessarily lead to high localization performance, as shown in Sec~\ref{sec:Results}. 
We follow the medium and hard protocols in ROxford and RParis~\cite{radenovic2018revisiting} with and without 1 \textit{M} distractors.

\noindent{\bf Pixel retrieval from ground-truth query-index image pairs.}
We can use existing ranking/reranking methods and treat the remaining process as one-shot detection/segmentation. In this case, the detection or segmentation performance is evaluated using the mean of mIoU of all the queries, where mIoU is the mean of the IoUs for all the ground-truth index images. We do not consider the false pairs because the ranking metric mAP well reflects the influence of false pairs in the ranking list.
\section{Applications of pixel retrieval}
\label{sec:Applications}
Pixel retrieval requires the machine to recognize, localize, and segment a particular first-seen object, which is one of the fundamental abilities of the human visual system.
It is useful for many applications. In this section, we first show that it can significantly improve the user experience in web search. We then discuss how pixel retrieval can help image-level ranking techniques. Finally, we introduce some other applications that may also benefit from pixel retrieval.

\noindent{\bf Web search user experience improvement.}
Modern image retrieval techniques focus on improving the image-level ranking performance of hard cases, such as images under extreme lighting conditions, novel views, or complicated occlusions. However, users may not easily perceive a hard case as a true positive, even if it is at the top of the ranking list. We claim that pixel-level annotation can significantly improve the user experience on the web search application. 

To see how pixel-level annotation improves the user experience on image search, we ran a user study where users were asked to find images that contain a given target among candidate images in two different conditions; the one with pixel-level annotations (\textit{i.e.,} Pixel retrieval) and the other with no annotations (\textit{i.e.,} Image retrieval). We recruited 40 participants on Prolific\footnote{prolific.co} and compared the time taken to complete the task between the two conditions. 

\begin{figure}[t]
    \centering
    \includegraphics[scale=0.4]{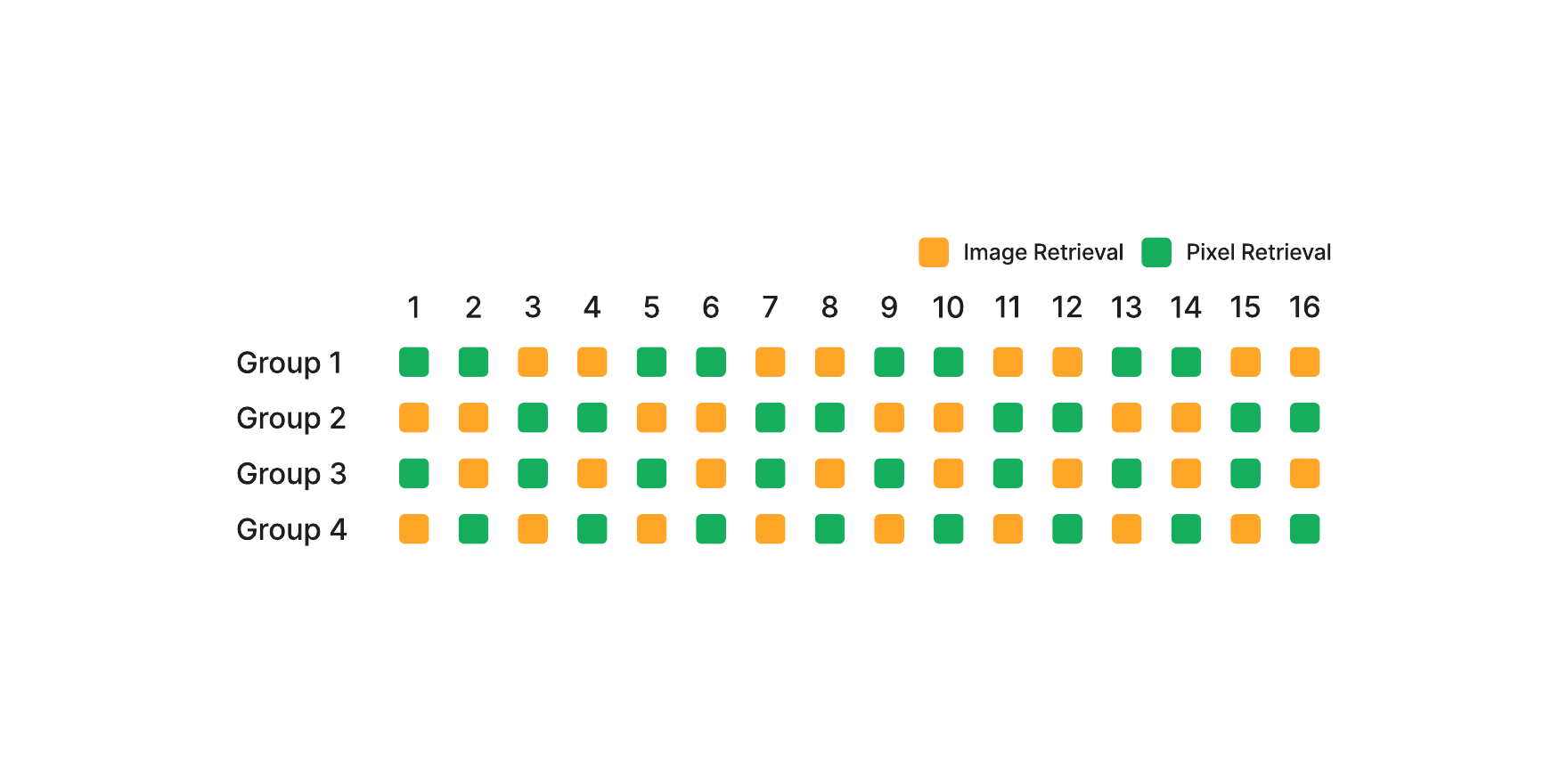}
    \caption{Design of the study on web search user experience. Image retrieval refers to a setting where no annotations are provided, whereas Pixel retrieval refers to a setting where pixel-level annotations are provided. 40 participants were divided into four groups and we counterbalanced the type of questions across the groups. Numbers 1 to 16 indicate the 16 questions.}
    \label{fig:study_design}
\end{figure}

Participants were asked to complete 16 questions in total, where eight of them were Pixel retrieval and the other eight were Image retrieval. We divided the participants into four groups and counterbalanced the type of questions (Figure~\ref{fig:study_design}). For each question, participants were given a query image and 12 candidate images. There were three true positives and nine false positives in the candidate images, and we randomly chose ground truth images of other queries as false positives. We shuffled the order of the candidate images and asked participants to choose three images that contain the query image (\textit{i.e.,} true positives) among them. Figure~\ref{fig:user_study} shows one of the 16 questions. You can check our user study from \href{https://fascinating-marzipan-a99b4c.netlify.app/bwds}{\textcolor{red}{this link}}.
To start the user study, please enter any character into the ``unique Prolific ID'' blank. Anonymity is guaranteed. 

Our results show that participants completed the task faster when the pixel-level annotations were presented (mean=37.07s, std=49.76s) than when no annotations were presented (mean=53.71s, std=80.08s). The difference between two conditions is statistically significant (T-test, p-value=0.00091), and participants responded that it was helpful to see annotations in completing the task  (mean=6.375/7, std=0.89).

\noindent{\bf Other applications.}
Image retrieval techniques have been applied to many applications, such as medical diagnosis and geographical information systems (GIS).
The pixel-level retrieval is also desirable for these applications. For example, the size of medical and geographical images are usually huge, and the doctors and GIS experts are interested in retrieving regions of the particular structures or landmarks from the whole images in the database~\cite{yang2014parallel,breznik2022cross,mehta2009content,zhu2021vigor}. 

Pixel retrieval can also help image matting~\cite{wei2021improved,xu2017deep,zhang2019late}. Current image matting techniques rely on the user's click to confirm the target matting region~\cite{wei2021improved,xu2017deep,zhang2019late}. Our pixel retrieval provides a new interaction method: deciding the target object based on the query example. This query-based interaction can significantly reduce user effort in situations where many images depict the same object~\cite{shen2016regional}. 

\section{Experiment}
\label{sec:Experiment}

We evaluate the performance of state-of-the-art (SOTA) methods in multiple fields on our new pixel retrieval benchmarks. Our new pixel retrieval task is a visual object recognition problem. It requires the search engine to automate the human visual system's ability to identify, localize, and segment an object under illumination and viewpoint changes. It can be seen as a combination of image retrieval, one-shot detection, and one-shot segmentation. We introduce these related tasks and their SOTA methods in this section, and we implement these SOTA methods and discuss their results in Section~\ref{sec:Results}. 

\subsection{Localization in retrieval}
Some pioneering works~\cite{lampert2009detecting,lin2010local,shen2013spatially} in image retrieval emphasized the importance of localization and tried to combine the retrieval and detection methods. However, due to the lack of a standard pixel retrieval benchmark, these pioneering works only showed qualitative results instead of quantitative comparisons. In this paper, we implement and compare the SOTA localization-related retrieval methods on our new benchmark dataset. They can be divided into two categories: spatial verification (SP) and detection. 

SP~\cite{shen2013spatially,lowe2004distinctive,arandjelovic2012three,Noh_2017_ICCV,cao2020unifying} is one of the most popular reranking approaches in image retrieval. It is also known as image matching~\cite{Jin2020}; SP and stereo task in Image Matching Challenge (IMC)~\cite{Jin2020} share the same pipeline and theory except for the final evaluation step. In this work, we selected the local features and matching hyperparameters with the best retrieval performance on ROxford and RParis, which contain more challenging cases than datasets in IMC. 

SP compares the spatial configurations of the visual words in two images. Theoretically, it can achieve verification and localization simultaneously. However, the image-level ranking performance cannot fully reflect the SP accuracy or localization performance. In the hard positive cases, \textit{e.g.}, where many repeated patterns exist in the background, even though SP generates a high inlier number and ranks an image on top of the ranking list, the matched visual words can be wrong due to the repeated patterns. Our pixel retrieval benchmark can not only evaluate the localization performance, but also better reflect the SP accuracy and be helpful for future SP studies.  

Researchers mainly focus on generating better local features to improve SP performance. The classical local features have SIFT~\cite{lowe2004distinctive}, SUFT~\cite{bay2006surf}, and rootSIFT~\cite{arandjelovic2012three}. Recently, DELF~\cite{Noh_2017_ICCV} and DELG~\cite{cao2020unifying} local features, which are learned from the large landmarks training set~\cite{weyand2020google}, achieve the SOTA SP result. We evaluate the SP performance with SIFT, DELF, and DELG features on our new benchmark datasets in this paper. 

Another localization-related image search approach is to directly apply the detection methods~\cite{lampert2009detecting,lin2010local,razavian2016visual,ren2015faster,uijlings2013selective,tao2014locality,teichmann2019detect}. Faster-RCNN~\cite{ren2015faster} and SSD detector~\cite{liu2016ssd} fine-tuned on a huge manually boxed landmark dataset~\cite{teichmann2019detect} achieve the SOTA detect-related retrieval result~\cite{teichmann2019detect}. Detect-to-retrieve (D2R)~\cite{teichmann2019detect} uses these fine-tuned models to detect several landmark regions for a database image and uses aggregation methods like the Vector of Locally Aggregated Descriptors (VLAD)~\cite{jegou2010aggregating} and the Aggregated Selective Match Kernel (ASMK)~\cite{tolias2013aggregate} to represent each region. To better check the effect of the aggregation methods, we also implement the Mean aggregation (Mean), which simply represents each region using the mean of its local descriptors. The region with highest similarity can be seen as the target region for a given query. We evaluate the combination of different detectors and aggregation methods on our pixel retrieval benchmarks.

\subsection{One-shot detection and segmentation}
We can treat pixel retrieval as combining image retrieval and one-shot detection and segmentation. We test the performance of these approaches. 


The Vision Transformer for Open-World Localization (OWL-ViT)~\cite{minderer2022simple} is a vision transformer model trained on the large-scale 3.6 billion images in LiT dataset~\cite{zhai2022lit}. It has shown the SOTA performance on several tasks including one-shot detection. 
The One-Stage one-shot Detector (OS2D) combines and refines the traditional descriptor matching and spatial verification pipeline in image search to do the one-shot detection. It achieves impressive detection performance in several domains, \textit{e.g.}, retail products, buildings, and logos. We test these two detection methods on our new benchmarks.

The Hypercorrelation Squeeze Network (HSNet)~\cite{min2021hypercorrelation} is one of the most famous few-shot segmentation methods. It finds multi-level feature correlations for a new class. The Mining model (Mining)~\cite{yang2021mining} exploits the latent novel classes during the offline training stage to better tackle the new classes in the testing time. The Self-Support Prototype model (SSP)~\cite{fan2022ssp} generates the query prototype in testing time and uses the self-support matching to get the final segmentation mask. The self-support matching is based on one of the classical Gestalt principles~\cite{koffka2013principles}: pixels of the same object tend to be more similar than those of different objects. It achieves the SOTA few-shot segmentation results on multiple datasets. We evaluate these three methods on our new pixel retrieval benchmarks. 



\subsection{Dense matching}

Different from image matching (SP in this paper), which calculates the transformation between two images of the same object from different views, dense matching focuses on finding dense pixel correspondence. We check if we can use the SOTA dense matching methods to correctly find the correspondence points for pixels in the query image and achieve our pixel retrieval target. 

GLUNet~\cite{truong2020glu} and RANSAC-flow~\cite{shen2020ransac} are popular among many famous dense matching methods. Recently, Truong~\etal have shown that the warp consistency objective (WarpC)~\cite{truong2021warp} and the GOCor module~\cite{truong2020gocor} can further improve the performance and achieve the new SOTA. Another popular method is PDC-Net~\cite{truong2021pdc}. It can predict the uncertainty of the matching pixels. The uncertainty can be useful for our pixel retrieval task, which is sensitive to the outliers. We test the origin GLUNet, GLUNet with WarpC (WarpC-GLUNet), GLUNet with GOCor module (GOCor-GLUNet), and PDC-Net in Table~\ref{tab:miou}. 

\Skip{
 We mainly test the SOTA models trained on MegaDepth~\cite{li2018megadepth}. MegaDepth is a good baseline training set for our new benchmark. Its real scene images is from the same source with Oxford and Paris, \textit{i.e.}, Flickr. It uses SfM reconstruction to find the ground-truth pixel correspondences; the famous image search training set SfM~\cite{radenovic2018fine} also uses the SfM reconstruction to label the image pairs for images from Flickr. Besides MegaDepth, we also test the models trained on the semantic dataset~\cite{} to see if it can give any help. } 

\subsection{Experiment detail}

We try our best to find the best possible result for each method on our novel benchmark. 
The retrieval localization methods employed in this study, including image matching (SP in this paper) and D2R, were configured to achieve optimal performance on ROxford and RParis. These methods rely on precise localization to enhance image retrieval performance. Thus, we adopt the same experimental configurations in our similar pixel retrieval benchmark. 
\Skip{
The retrieval localization methods employed in this study, including image matching (SP in this paper) and D2R, were configured to achieve optimal performance on ROxford and RParis. These methods rely on precise localization to enhance image retrieval performance; hence, the optimal settings for image retrieval in the same datasets might also yield good outcomes for our pixel retrieval
\textcolor{green}{WJ: The retrieval localization methods employed in this study, including image matching (SP in this paper) and D2R, were configured to achieve optimal performance on ROxford and RParis. Thus, we adopt the same experimental configurations in our similar pixel retrieval benchmark.}}
Similarly, dense matching methods, which encompass geometric and semantic matching tasks, are expected to operate directly on our pixel retrieval benchmark, as per task definitions. We evaluate its geometric models with the best performance on MegaDepth~\cite{li2018megadepth} and ETH3D~\cite{schops2017multi}, datasets that feature actual building images, rendering them the ideal valid sets for our benchmark. The difference is that our dataset contains more extreme viewpoints and illumination changes. Moreover, we evaluate the performance of semantic models to see if including semantic information can enhance rigid body recognition in our benchmarks.
We refrained from fine-tuning the segmentation methods as there is no segmentation training set pertaining to the building domain to the best of our knowledge. 
Our comprehensive experimental findings can be employed as baseline metrics for future comparisons. 
We include the detailed experimental configurations for each method in the supplementary materials and intend to make them, along with their codes, publicly available.

\begin{table*}[htp]
    \caption{Results of pixel retrieval from ground truth query-index image pairs (\% mean of mIoU) on the PROxf/PRPar datasets with both Medium and Hard evaluation protocols. D and S indicate detection and segmentation results respectively. \textbf{Bold} number indicates the best performance in each field; \textbf{\textcolor{red}{red}} number indicates the best performance throughout all fields. }
    \label{tab:miou}
    \centering
    \begin{tabular}{|l | c | c | c | c | c | c | c | c|}
    \hline
   \multirow{3}{*}{Method} & \multicolumn{4}{c}{Medium} & \multicolumn{4}{|c|}{Hard} \\
    \cline{2-9}
     & \multicolumn{2}{|c|}{PROxf}  & \multicolumn{2}{|c|}{PRPar} & \multicolumn{2}{|c|}{PROxf} & \multicolumn{2}{|c|}{PRPar}  \\
     \cline{2-9}
          & D & S & D & S & D & S & D & S  \\
    \hline
    \multicolumn{9}{|c|}{Localization methods in retrieval}\\
    \hline
        SIFT+SP~\cite{philbin2007object} &10.5  &3.9 &14.0 &5.1 &7.1  &2.4  &12.4 &4.3 \\
        DELF+SP~\cite{Noh_2017_ICCV} &14.5  &\textbf{5.5}  &21.3 &\textbf{7.5} &9.4 &\textbf{4.1} &16.7 &\textbf{5.5}\\
        DELG+SP~\cite{cao2020unifying} &13.8 &5.2  &18.6 &7.2 &8.9 &2.9  &13.6 &4.9 \\
        D2R~\cite{teichmann2019detect}+Resnet-50-Faster-RCNN+Mean  &20.2 &-  &29.6 &- &16.7 &-  &27.4 &- \\
        D2R~\cite{teichmann2019detect}+Resnet-50-Faster-RCNN+VLAD~\cite{jegou2010aggregating} &25.8  &- &37.5 &- &\textbf{21.6}  &-  &35.5 &- \\
        D2R~\cite{teichmann2019detect}+Resnet-50-Faster-RCNN+ASMK~\cite{tolias2016image} &\textbf{26.3}  &-  &\textbf{38.5} &- &\textbf{21.6} &- &\textbf{\textcolor{red}{35.6}} &-\\
        D2R~\cite{teichmann2019detect}+Mobilenet-V2-SSD+Mean &19.7 &-  &25.9 & -&20.1 &-  &27.9 &- \\
        D2R~\cite{teichmann2019detect}+Mobilenet-V2-SSD+VLAD~\cite{jegou2010aggregating}  &23.1 &-  &33. &- &20.9 &-  &33.6 &- \\
        D2R~\cite{teichmann2019detect}+Mobilenet-V2-SSD+ASMK~\cite{tolias2016image}   &22.4 & - &34.0 &- &20.8 &-  &33.1 &- \\
    \hline
    \multicolumn{9}{|c|}{One-shot detection and segmentation methods}\\
    \hline
    OWL-VIT (LiT)~\cite{minderer2022simple} & 11.4 & - & 18.0 & - & 6.3 & - & 15.0 & - \\
    OS2D-v2-trained~\cite{osokin20os2d} & 10.5 &-  &13.7 &- &11.7 &- &14.3 &-\\
    OS2D-v1~\cite{osokin20os2d}  &7.0 &-  &8.5 &- & 8.7&-  &9.2 &- \\
    OS2D-v2-init~\cite{osokin20os2d}   &13.6 &-  &15.4 &- &14.0 &-  &15.1 & -\\

    SSP (COCO) + ResNet50~\cite{fan2022ssp} & 19.2 & 34.5 & 31.1 & 48.7 & 15.1 & 25.3 & 29.8 & \textbf{\textcolor{red}{41.7}} \\
     SSP (VOC) + ResNet50~\cite{fan2022ssp}  & 19.7 & 34.3 & 31.4 & \textbf{\textcolor{red}{48.8}} & 16.1 & \textbf{\textcolor{red}{26.1}} & 30.3 & 40.4 \\
     HSNet (COCO) + ResNet50~\cite{min2021hypercorrelation} & 23.4 & 32.8 & 37.4 & 41.9 & 21.0 & 25.7 & 34.7 & 36.5 \\
     HSNet (VOC) + ResNet50~\cite{min2021hypercorrelation} & 21.0 & 29.8 & 31.4 & 39.7 & 17.1 & 23.2 & 29.7 & 34.9 \\
     HSNet (FSS) + ResNet50~\cite{min2021hypercorrelation} & \textbf{30.5} & \textbf{\textcolor{red}{35.7}} & \textbf{\textcolor{red}{39.4}} & 40.2 & \textbf{\textcolor{red}{22.7}} & 25.1 & \textbf{34.7} & 32.8 \\
     Mining (VOC) + ResNet50~\cite{yang2021mining} & 18.3 & 30.5 & 29.6 & 42.7 & 15.1 & 21.4 & 28.1 & 34.3 \\
     Mining (VOC) + ResNet101~\cite{yang2021mining} & 18.1 & 28.6 & 29.5 & 40.0 & 14.2 & 20.4 & 28.2 & 34.4 \\
        \hline
    \multicolumn{9}{|c|}{Dense matching methods}\\
    \hline
    
    GLUNet-Geometric~\cite{truong2020glu} & 18.1 & 13.2 & 22.8 & 15.2 & 7.7 & 4.6 & 13.3 & 7.8 \\
    PDCNet-Geometric~\cite{truong2021pdc} & 29.1 & 24.0 & 30.7 & 21.9 & 20.4 & 15.7 & 20.6 & 12.6 \\
    GOCor-GLUNet-Geometric~\cite{truong2020gocor} & 30.4 & \textbf{26.0} & 33.4 & 25.6 & 20.8 & \textbf{16.0} & 19.8 & 13.3 \\
    WarpC-GLUNet-Geometric (megadepth)~\cite{truong2021warp} & \textbf{\textcolor{red}{31.3}} & 25.4 & 36.6 & \textbf{27.3} & \textbf{21.9} & 15.8& 26.4 & 17.3 \\
    WarpC-GLUNet-Geometric (megadepth$\_$stage1)~\cite{truong2021warp} & 23.5 & 19.3 & 28.1 & 20.7 & 13.2 & 8.9 & 17.0 & 10.9 \\
    GLUNet-Semantic~\cite{truong2020glu} & 18.5 & 14.4 & 22.4 & 15.6 & 8.7 & 5.6 & 12.8 & 7.8 \\
    WarpC-GLUNet-Semantic~\cite{truong2021warp} & 27.5 & 21.4 & \textbf{36.8} & 25.7 & 18.5 & 11.9 & \textbf{28.3} & \textbf{17.6} \\
    
    \hline

    \end{tabular}

\end{table*}

\section{Results and discussion}
\label{sec:Results}

\begin{figure*}[t]
    \centering
    \includegraphics[scale=0.12]{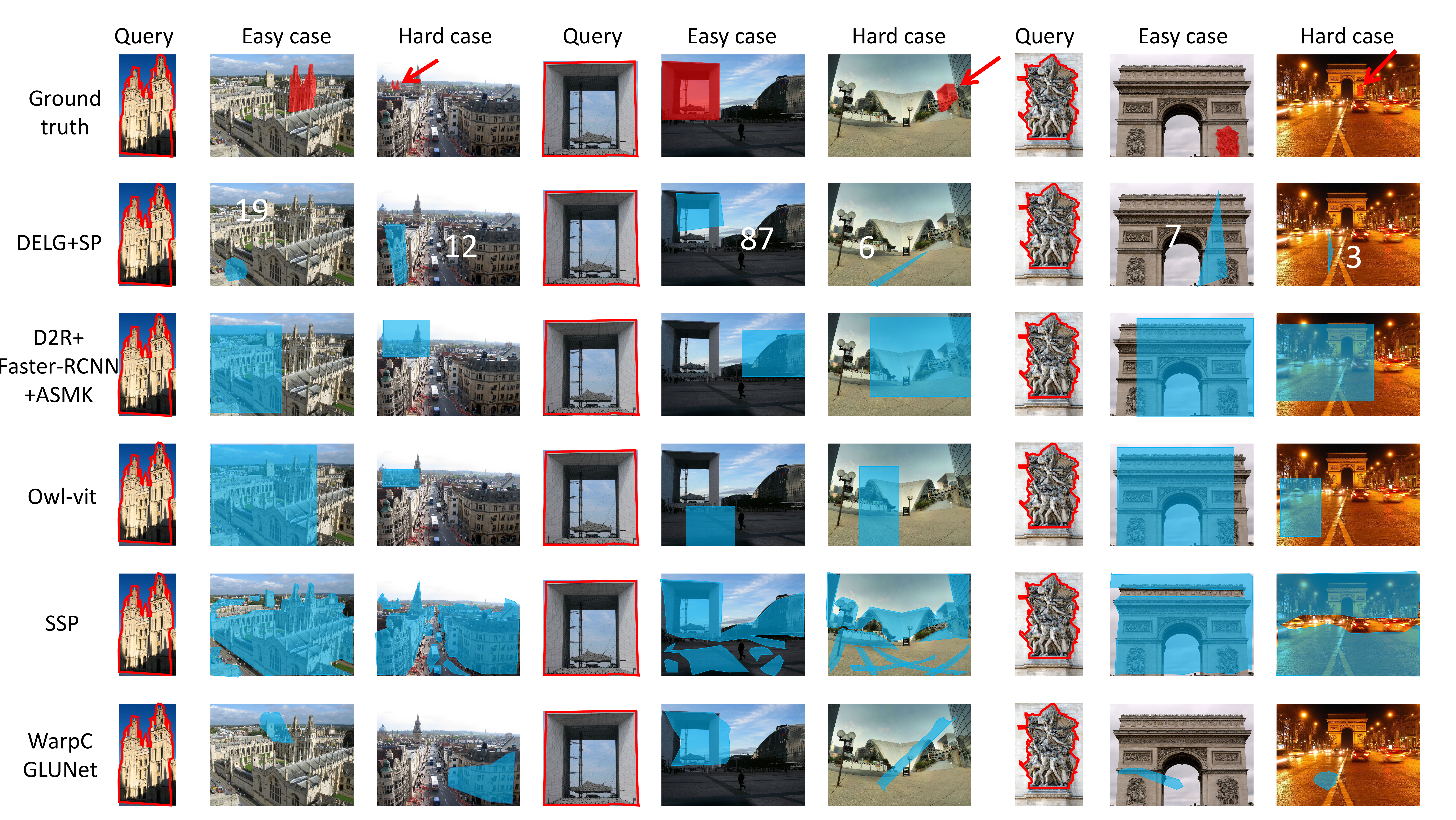}
    \caption{Qualitative comparison of the SOTA methods in different fields on the pixel retrieval benchmarks. 
    Blue masks represent the prediction results of each method. 
    For SP and WarpCGLUNet, we consider the union of all the matching points as the prediction masks.
    We also show the inlier numbers for the SP method. Pixel retrieval is challenging for existing methods and further research is needed.
    }
    \label{fig:qualitative}
\end{figure*}

\begin{table*}[hbt]
    \caption{Results of pixel retrieval from database (\% mean of mAP@50:5:95) on the PROxf/PRPar datasets and their large-scale versions PROxf+1M/PRPar+1M, with both Medium (M) and Hard (H) evaluation protocols. \textbf{Bold} indicates the best performance using the same image ranking list; \textbf{\textcolor{red}{red}} indicates the best performance in two ranking lists. \colorbox{green}{Green} lines show the image level mAPs of the ranking lists.}
    \label{tab:map}
    \centering
    \begin{tabular}{|c| c | c| c| c| c| c| c| c| c| }
    \hline
    \multicolumn{2}{|c|}{\multirow{2}{*}{}} & \multicolumn{2}{c|}{PROxf} & \multicolumn{2}{c|}{PROxf+R1M} & \multicolumn{2}{c|}{PRPar} & \multicolumn{2}{c|}{PRPar+R1M}   \\ \hline
         \multicolumn{2}{|l|}{}& M & H & M & H & M & H & M & H  \\ \hline 
         \multicolumn{10}{|c|}{Image retrieval: DELG initial ranking~\cite{cao2020unifying}} \\ \hline
         \rowcolor{green} \multicolumn{2}{|c|}{Image level mAP} &76.3  &55.6  &63.7 &37.5 &86.6  &72.4  &70.6 &46.9 \\ \hline
         \multirow{5}{*}{ \makecell[c]{Pixel \\retrieval \\methods}}&
         DELG + SP~\cite{cao2020unifying} &6.1  &6.3  &5.8 &6.7 &10.9  &8.0  &10.5 &7.8 \\
         &D2R+Faster-RCNN+ASMK~\cite{teichmann2019detect} &29.6  &22.5  &28.8 &19.1 &26.3  &25.6  &23.7 &20.5\\
         &OWL-VIT~\cite{minderer2022simple} & 13.1 &8.1  &12.8 &7.2 &8.3  &12.7  &7.6 &11.4\\
         &SSP~\cite{fan2022ssp} &\textbf{\textcolor{red}{37.3}}  &34.6  &\textbf{\textcolor{red}{36.6}} &29.9 &\textbf{\textcolor{red}{47.0}}  &\textbf{\textcolor{red}{43.1}}  &\textbf{44.5} &\textbf{37.1}\\
         &WarpCGLUNet~\cite{truong2021warp} &34.3  &\textbf{\textcolor{red}{36.8}}  &33.9 &\textbf{\textcolor{red}{34.9}} &33.9  &28.8  &32.9 &27.1\\
        
        \hline
        \multicolumn{10}{|c|}{Image retrieval: DELG initial ranking~\cite{cao2020unifying} + HP reranking~\cite{an2021hypergraph}} \\ \hline
        \rowcolor{green} \multicolumn{2}{|c|}{Image level mAP} &85.7  &70.3  &78.0 &60.0 &92.6 &83.3  &86.6 &72.7 \\ \hline
        \multirow{5}{*}{ \makecell[c]{Pixel \\retrieval \\methods}}& DELG + SP~\cite{cao2020unifying} & 6.4 &7.2  &6.2 &7.5 &10.7  &6.0  &10.7 &5.9 \\
         &D2R+Faster-RCNN+ASMK~\cite{teichmann2019detect} &30.1  &23.5 &30.5 &22.0 &26.3  &25.3  &25.7 &24.9\\
         &OWL-VIT~\cite{minderer2022simple}  &12.3  &6.6 &12.1 &13.6 &7.9  & 7.6 &7.9 &7.8\\
         &SSP~\cite{fan2022ssp}  & \textbf{33.0} &29.7  &\textbf{35.7} &\textbf{30.5} & \textbf{46.4} &\textbf{37.2}  &\textbf{\textcolor{red}{45.6}} &\textbf{\textcolor{red}{37.2}}\\
         &WarpCGLUNet~\cite{truong2021warp}  & 31.2 &\textbf{32.6}  &31.5 &31.7 &34.1  &27.3  & 34.3&28.1\\ 
         \hline
    
    \end{tabular}

\end{table*}

We report the results of pixel retrieval from ground-truth image pairs (mean of mIoU) for all the above mentioned methods in Table~\ref{tab:miou}. We choose one to two representative methods for each field and show their qualitative results in Figure~\ref{fig:qualitative}. 
To evaluate the performance of pixel retrieval from database, we combine these methods with SOTA image level ranking and reranking methods: DELG and hypergraph propagation (HP)~\cite{an2021hypergraph}. We show their final mAP@50:5:95 in Table~\ref{tab:map}.

Although SP  
achieves impressive image-level retrieval results~\cite{cao2020unifying, Noh_2017_ICCV}, it shows suboptimal performance on pixel retrieval. We observe some true positive pairs where SP gives a high inlier number but matches the wrong regions. For example, in the first easy case in Figure~\ref{fig:qualitative}, SP with DELG features generates 19 inliers, but none of the inliers are in the target object region. Note that 19 inlier number is high and only 4 false positive images are ahead of the this easy case in the final DELG reranking list~\cite{cao2020unifying}. This is not to say DELG is bad; in fact, its matching results are quite good in most cases. We choose this striking example only to show that the image-level ranking performance is not enough to reflect the SP accuracy. Our pixel retrieval benchmarks can be used to evaluate the matched features' locations of SP. 

For SP, both deep-learning features DELF and DELG significantly outperform the SIFT features. Interestingly, although DELG shows better image retrieval performance~\cite{cao2020unifying} than DELF, it is slightly inferior to DELF in the pixel retrieval task. One reason might be that though DELG generates more matching inliers for the positive pairs than DELF, these inliers tend to exist in a small region and do not reflect the location or size of the target object. Improving SP performance in both image and pixel level can be a practical research topic.  

Although the detect-2-retrieval~\cite{teichmann2019detect} is inferior to SP in image retrieval~\cite{cao2020unifying,Noh_2017_ICCV,radenovic2018revisiting}, it shows better performance than SP in our pixel-level retrieval benchmarks. We conjecture that the detection models tend to cover the whole building more than SP. Our benchmark is helpful in checking this conjecture and designing a better pixel retrieval model for future works. The results of the region detector and the aggregation method are similar to the trend in image search~\cite{teichmann2019detect}. The VLAD and ASMK aggregation methods significantly improve the Mean aggregation. A faster-RCNN-based detector shows better performance than SSD. 

For dense matching methods, GLU-Net using warp consistency or GOCor module and PDC-Net show better results than other models. This trend is similar to that in the dense matching benchmark Megadepth~\cite{li2018megadepth}. 

The segmentation methods significantly outperform other methods in terms of the mean of segmentation mIoU.
However, their detection mIoU results are not so impressive. They tend to predict the entire foreground, which contains the target building, as shown in the SSP line of Figure~\ref{fig:qualitative}. Among the segmentation methods, SSP shows better segmentation than others, showing its self-support approach is helpful for finding more related pixels.  




Another interesting finding is that better image ranking mAP does not necessarily brings better pixel retrieval  mAP@50:5:95, as shown in Table~\ref{tab:map}. The reason might be that the image search techniques rank some hard cases high, but detection methods do not well localize the query object in them.

\Skip{
Surprisingly, segmentation and dense matching methods show better mIoU results than matching and localization methods in the image search field, even though they are not designed for retrieval tasks initially. However, they must work with image search techniques to tackle our pixel retrieval task. Although the existing dense matching and segmentation methods detect as much as possible the target object areas, they are not designed to achieve fine-grained recognition. On the contrary, existing retrieval methods tend to identify certain textures or corners but are not designed to find the whole object's shape. Without a proper benchmark, retrieval methods can mismatch an object and its context to improve image-level performance, as we discussed above. Our new benchmark, which is difficult and has careful quality checking, can provide a good evaluation for new methods targeting the fine-grained detection and segmentation problem.}

It is interesting to note that segmentation and dense matching methods have demonstrated superior mIoU results compared to matching-based and detection-based retrieval methods, despite not being originally designed for retrieval tasks. However, to effectively tackle the pixel retrieval task, these methods must work in conjunction with image search techniques. While dense matching and segmentation methods are better suited for identifying target object areas, they may not achieve fine-grained recognition. In contrast, existing retrieval methods tend to identify certain textures or corners but lack the ability to capture the entire object's shape. Without a reliable benchmark, retrieval methods may simply associate an object and its context to improve image-level performance, leading to low localization and segmentation results, as we discussed above. We did our best to prepare our new benchmark so that it can provide a valuable evaluation for novel methods targeting pixel retrieval, which requires fine-grained and variable-granularity detection and segmentation.
Moreover, we find pixel retrieval challenging. The current best mAP@50:5:95 in PROxford and PRParis at medium setting without distractors are only 37.3 and 47.0.

\vspace{-5pt}

\section{Future works}

We present a novel task termed "Pixel Retrieval." This task mandates segmentation but transitions from a semantic directive to the content-based one, thus bypassing semantic vagueness. Concurrently, it demands large-scale, instance-level recognition—a subject frequently explored by the retrieval community. This innovative task poses several unique challenges, some of which we outline below:

\subsection{Enhancing accuracy}

For a superior user experience, it's vital to embrace methods, workflows, and datasets that bolster accuracy. Our findings illustrate that segmentation and dense matching methods are beneficial, especially when an image ranking list is provided using existing retrieval techniques. Beyond merely superimposing segmentation over retrieval, a compelling approach would be to rank images based on the results of the segmentation. Further insights and experimental outcomes in this regard are available on our website.

Although the introduction of new datasets, even those echoing the landmarks in our benchmarks, is commendable, it's pivotal to articulate their application to discern the sources of performance enhancements. If PROxford/PRParis and ROxford/RParis are employed as benchmarks, it's crucial to ensure the consistent usage of the same training set. Given the public accessibility of our ground truth files, it's imperative to prevent any unintended data leaks during training.

\subsection{Scalability and speed}

A major challenge lies in scaling the algorithms and augmenting the retrieval speed. Techniques like segmentation and dense matching, which compute for every pair, inherently lag in speed when compared to retrieval methods such as ASMK and D2R. Therefore, swift methods that can cater to extensive scales are highly sought after.

\subsection{Innate visual recognition and The significance of training data}

The prevalent trend in research is to amass expansive training or fine-tuning sets closely aligned with test instances—certainly a commendable approach. However, intriguingly, humans exhibit an innate ability to discern instances in query images. Our annotators, despite being unfamiliar with European landmarks, could effortlessly segment target objects in each positive image, even when subjected to extreme lighting and perspective alterations. What fuels this innate recognition? Is it purely due to extensive prior exposure, or are there underlying mechanisms at play? How pivotal is the training dataset in replicating human-like content-based segmentation, especially when semantic influences are excluded? These questions beckon exploration.


\section{Conclusion}

We introduced the first landmark pixel retrieval benchmark datasets, \textit{i.e.,} PROxford and PRParis, in this paper. To create these benchmarks, three professional annotators labeled, refined, and checked the segmentation masks for a total of 5,942 image pairs. We executed the user study and found that pixel-level annotation can significantly improve the user experience on web search; pixel retrieval is a practical task. We did extensive experiments to evaluate the performance of SOTA methods in multiple fields on our pixel retrieval task, including image search, detection, segmentation, and dense matching. Our experiment results show that pixel retrieval is challenging and further research is needed.

{\small
\bibliographystyle{ieee_fullname}
\bibliography{egpaper_for_review}
}

\end{document}